\documentclass[conference]{IEEEtran}
\IEEEoverridecommandlockouts
\usepackage{amsmath,amssymb,amsfonts,amsthm}
\usepackage{algorithmic}
\usepackage{graphicx}
\usepackage{subcaption}
\usepackage{hyperref}
\usepackage{array}
\usepackage{float}
\usepackage{caption}
\usepackage[ruled]{algorithm2e}  
\usepackage{textcomp}
\usepackage{xcolor}
\def\BibTeX{{\rm B\kern-.05em{\sc i\kern-.025em b}\kern-.08em
    T\kern-.1667em\lower.7ex\hbox{E}\kern-.125emX}}

\newtheorem{definition}{Definition}

\begin{document}

\title{Incremental Learning of Affordances using Markov Logic Networks\\
}

\author{\IEEEauthorblockN{1\textsuperscript{st} George B.G. Potter}
\IEEEauthorblockA{\textit{TNO}\\
Den Haag, The Netherlands \\
george.potter@tno.nl}
\and
\IEEEauthorblockN{2\textsuperscript{nd} Gertjan Burghouts}
\IEEEauthorblockA{\textit{TNO}\\
Den Haag, The Netherlands \\
gertjan.burghouts@tno.nl}
\and
\IEEEauthorblockN{3\textsuperscript{rd} Joris Sijs}
\IEEEauthorblockA{\textit{Cognitive Robotics} \\
\textit{TU Delft}\\
Delft, The Netherlands \\
j.sijs@tudelft.nl}
}

\maketitle

\begin{abstract}
Affordances enable robots to have a semantic understanding of their surroundings. This allows them to have more acting flexibility when completing a given task. Capturing object affordances in a machine learning model is a difficult task, because of their dependence on contextual information. Markov Logic Networks (MLN) combine probabilistic reasoning with logic that is able to capture such context. Mobile robots operate in partially known environments wherein unseen object affordances can be observed. This new information must be incorporated into the existing knowledge, without having to retrain the MLN from scratch. We introduce the MLN Cumulative Learning Algorithm (MLN-CLA). MLN-CLA learns new relations in various knowledge domains by retaining knowledge and only updating the changed knowledge, for which the MLN is retrained. We show that MLN-CLA is effective for accumulative learning and zero-shot affordance inference, outperforming strong baselines.
\end{abstract}


\section{Introduction}

Affordances play an important role in semantic understanding of scenes in robotics. These affordances, first introduced by Gibson \cite{gibson1979}, are the potential actions that an object affords to an agent depending on object properties and state, action effects, situational context and agent capabilities. In a robotics context, affordances model the relation between the robot, an object, and the possible interactions between the two \cite{mihai2018aff-eq}.
These affordances allow the robot to reason about its beliefs of the world in relation to the tasks and actions it may execute within the environment. Particularly in partially known environments, these affordances, in combination with reasoning about them, may result in more options for the robot to choose from. As a result affordances increase the possibility of the robot successfully completing its task 
\cite{ardon2020affordances}.
%
The robot can fuse this sensor data to reason about action possibilities present in the environment. 
Including a component in the perception-action system to reason about affordances, also called affordance inference, would result in a robot that is more effective in planning its task given a current belief of the surrounding world.

Reasoning about affordances is strongly related to symbolic reasoning as an affordance is a symbolic concept. In symbolic reasoning, raw data is abstracted into symbols representing the data, such as \texttt{Heavy} representing a range of masses or \texttt{Green} representing a certain range of RGB colours. The semantic meaning of these symbols is easy to understand for an operator. Symbols can be chained and fit together in a syntax to form a language that machines can use to reason about, and people can use to read. Some languages allow for logic to be applied to symbols. This is called symbolic reasoning with logic. Robots that have an understanding of symbols and logic can use the language and reasoning to build their knowledge bases.
%
%
An example of a (mathematical) language that embeds symbolic knowledge in logic is first-order logic (FOL). 
Challenges are contradicting formulas and the Boolean interpretation of logic. Markov Logic Networks can solve these problems \cite{richardson2006mln, domingos2019unifying}. 

A Markov Logic Network (MLN) is a knowledge base of first-order logic formulas with a weight attached to each formula \cite{richardson2006mln, domingos2019unifying}. MLNs can compactly represent regularities in the world and allow reasoning over these regularities. The weight of a formula in the knowledge base is a measure of how likely that formula is to occur given observations of the world. \autoref{tab:mln-example} provides an example MLN that consists of three formulas. The formulas do not conflict logically, but semantically seem incorrect when taking into account that each formula is $\forall x, y$. This is reflected in the formula weights that soften the $\forall$ constraints. The third formula has a negative weight, indicating that the formula is false most of the time. 


\begin{table}[t]
    \centering
    \caption{A Markov Logic Network consisting of three formulas. Each formula has a weight to indicate how often a formula is true relative to the other formulas in the MLN.}
    \label{tab:mln-example}
    \vspace{-0.1cm}
    \begin{tabular}{c|l}
        \textbf{Weight} & \textbf{Formula}\\ \hline
         1.3 &  $\texttt{SharpEdge(x)} \Rightarrow \texttt{Affordance(x, Cutting)}$\\
         2.8 &  $\texttt{Heavy(x)} \Rightarrow \texttt{!Affordance(x, Lifting)}$\\
         -4.5 & $\texttt{Affordance(x)} \Leftrightarrow \texttt{Affordance(y)}$\\
    \end{tabular}
    \vspace{-0.65cm}
\end{table}

In most applications MLN models are trained once before deployment. 
in a process called \textit{batch} or \textit{offline} learning \cite{benDavid1997offline}. Updating these models afterwards with newly obtained information from the real world requires retraining the MLN from scratch. It is thus not very efficient from a computational point of view as old data needs to be processed again. Updating only parts of the model relevant to the newly obtained information saves on compute resources and time \cite{cui2022mcla}. The Markov properties of MLNs enable partially updating the network \cite{richardson2006mln}. Newly obtained information can thus be integrated into the existing knowledge over time in a manner called \textit{cumulative}, \textit{online}, \textit{incremental} or \textit{lifelong} learning \cite{thórisson2019cumulative}. The task of affordance inference is suitable for incremental learning as object affordances are difficult to define a priori without additional context from a given situation. 
%
The key is to mitigate \textit{catastrophic forgetting}, i.e. overwriting (parts of) old knowledge during the learning of new knowledge, thus 
to potentially critical information necessary to complete a task \cite{french1999catastrophic}. 
To our knowledge, only two methods exist in the literature that extend Markov Logic Networks with cumulative learning capabilities: OSL and MCLA. The Online Structure Learning (OSL) \cite{huynh2011osl} method learns new formulas and new weights from incoming evidence at each time step. This method is prone to catastrophic forgetting due to the lack of a knowledge manager that prevents overwriting of old knowledge. 

\begin{figure}
    \centering
    \begin{subfigure}{0.4\textwidth}
        \centering
        \includegraphics[width=\textwidth]{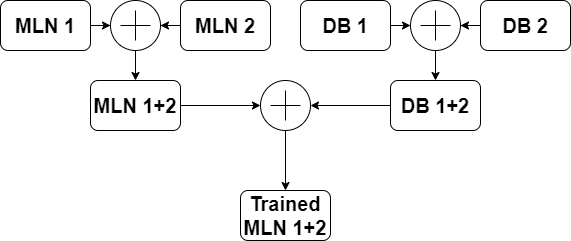}
        \caption{Batch learning}
        \label{fig:simple-batch}
    \end{subfigure}\\
    \vspace{0.35cm}
    \begin{subfigure}{0.4\textwidth}
        \centering
        \includegraphics[width=\textwidth]{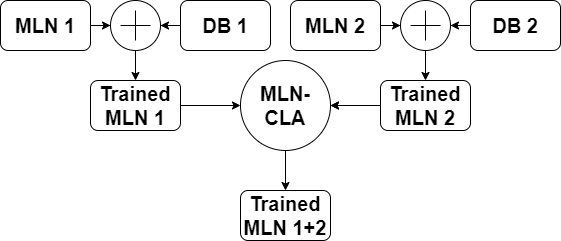}
        \caption{MLN-CLA (ours)}
        \label{fig:simple-mln-cla}
    \end{subfigure}
    \caption{In batch learning, two MLN-database pairs are combined into one before training. In MLN-CLA, each MLN-database pair is trained separately before merging.}\vspace{-0.5cm}
    \label{fig:batch-vs-ours}
\end{figure}

\begin{figure*}[t]
    \centering
    \includegraphics[width=\linewidth]{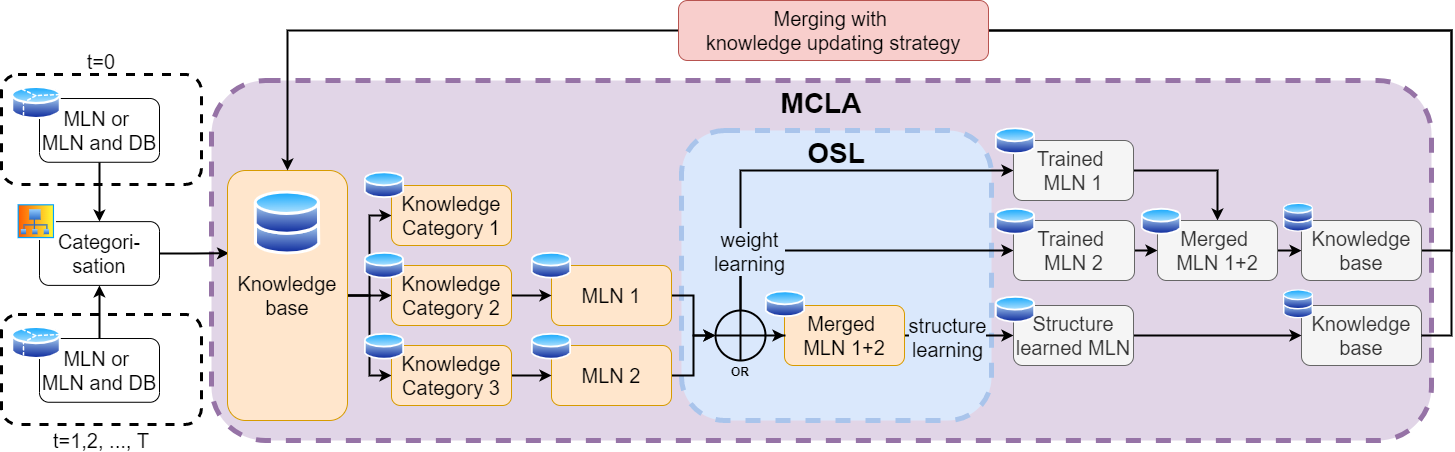}
    \caption{MLN-CLA: We build our knowledge categorisation and knowledge updating strategies on top of the same concepts as MCLA \cite{cui2022mcla}, outlined in purple. With the blue outlined box we indicate the overlap with OSL \cite{huynh2011osl}.}
    \label{fig:mln-cla-overview}
\end{figure*}

Our method mitigates catastrophic forgetting by introducing knowledge updating strategies that define conditions for which formula weight overwriting is preferred. The MLN Cumulative Learning Algorithm (MCLA) \cite{cui2022mcla} method employs a knowledge manager to cluster the knowledge in the knowledge base. The clustering helps in efficiently updating the formula weights of only relevant formulas in relation to newly obtained evidence. However, in MCLA all information about which formulas belong to which cluster is known a priori. Our approach introduces a knowledge clustering method based on co-occurrence of predicate domains in formulas. Our strategy is outlined in \autoref{fig:batch-vs-ours}. The cumulative learner introduced in this article, called MLN-Cumulative Learning Algorithm (MLN-CLA), will resolve these limitations of catastrophic forgetting and a-priori clustering of formulas. MLN-CLA enables a robot to reason over existing object affordances and learn new ones. An overview of our algorithm is given in \autoref{fig:mln-cla-overview}. We demonstrate a cumulative learner capable of learning new affordances, objects, object properties, and the multitude of relations between these concepts. 

\section{Related Work}\label{sec:rel-work}

Bayesian networks (BN) were used to represent relations between objects and affordances 
\cite{montesano2007,cabrera2019bayesian}.
%
Montesano et al. used a Bayesian network to model the interaction between objects, actions and effects \cite{montesano2007}. Their model learns many of these types of relations from observations of interactions with the environment. Based on a set of visual object properties such as size and shape, the model can infer possible actions and their effects for objects outside the set of objects the BN was trained on. Additionally, it can update its beliefs, i.e. the strength of a relation in the network, based on observations of interactions with objects. It can adjust its predictions over time as it learns from new object interactions. 
%
%
A major drawback of BNs is their graph structure and dependency representation. BNs cannot handle cyclic dependencies without additional extensions. For affordance inference, this is problematic: actions cause effects, effects result in changes in object state and object states influence the action space. 

Markov networks are similar to Bayesian networks but with the benefit that they do allow modelling of cyclic dependencies in an undirected graph. Moreover, Markov logic is a first-order logic that balances human interpretability with model expressiveness \cite{domingos2019unifying}.
%
A typical MLN cannot be updated over time, i.e. it is not ready for cumulative learning. To enable this property one must transform the MLN into a knowledge base wherein the formulas, their weights, variables and evidence are managed. Cui et al. introduced a `knowledge identifier' that is assigned a priori to manage the knowledge base \cite{cui2022mcla}. Their knowledge manager assigns a unique identifier to different clusters of knowledge.
Incoming evidence is categorised into these clusters, called Knowledge Categories, based on this knowledge identifier. To manage these clusters they introduce a Knowledge List. This Knowledge List is essentially a container for all Knowledge Categories. In MLN-CLA we adopt these concepts of Knowledge Categories and Knowledge Lists. We lift the requirement of labelling data beforehand through automatic categorisation. 
In contrast to \cite{cui2022mcla}, we show how to merge Knowledge Categories together to more efficiently cluster formulas in the Knowledge List.
%
Cui et al. do not manage overwriting essential old knowledge by newly learned knowledge. OSL \cite{huynh2011osl} has no knowledge manager. We introduce an explicit knowledge merging step that handles formula weight overwriting conflicts. These conflicts arising from a newer version of a formula overwriting the old formula. Depending on the chosen knowledge updating strategy the conflict is resolved and the user is in control of how catastrophic forgetting is handled. The strategies can be as simple as 'only overwrite if the new formula is based on more evidence than the old one' or as complex as the user wants. A new strategy can easily be defined and used within our framework.

\vspace{1cm} 
\section{Method}\label{sec:method}


Markov logic is a form of first-order logic (FOL). In FOL, formulas consist of four types of symbols: constants, variables, functions and predicates. Constants  represent things within a domain, such as objects (domain): \texttt{Ball} (constant), \texttt{Chair} or actions: \texttt{Throw}, \texttt{Kick}. Variables range over constants in a domain, e.g. \texttt{y}. Predicates represent relations over their arguments, e.g. \texttt{HasWeight(Ball, Light)} where \texttt{Ball} and \texttt{Light} are of the domain object and weight respectively. Functions are not considered in MLNs. With the constant, variable and predicate symbols any FOL formula can be constructed: $\forall y \; \texttt{HasWeigth(y, Heavy)} \Rightarrow \neg\texttt{Affordance(y, Throw)}$, where \texttt{y} is of the domain \texttt{object}. A world is defined by the truth values of all possible predicates and constants combinations.
%
Markov logic adds probabilistic weights to first-order logic formulas. These weights acts as soft constraints on the universal quantifier $\forall$ in FOL formulas. The weights indicate that a world that violates a formula is less probable but not impossible. A weight determines how strong the soft constraint is. An infinite weight is equal to first-order logic. A Markov Logic Network is a set of weighted first-order logic formulas. These formulas in combination with a set of constants form a Markov network \cite{pearl1988bp}. A Markov network is a graphical representation of the probability distribution of the possible worlds. MLNs are templates for the construction of Markov networks.
%
In MLN-CLA, illustrated in \autoref{fig:mln-cla-overview}, knowledge exists in the forms of Knowledge Lists, Knowledge Categories, MLN models and evidence. MLN models are simply the formulas with associated weights, such as in \autoref{tab:mln-example}, that capture knowledge about the world. Evidence is the actual data consisting of solely predicates of which all arguments are constants. This evidence is used to determine the weights of the formulas in the MLN model during training. The Knowledge List and Categories are further examined as these concepts form the core of knowledge management in MLN-CLA. Both concepts are adapted from Cui et al. \cite{cui2022mcla} and further refined.


\subsection{Updating Knowledge Lists}
A Knowledge List $\mathcal{L}$ consists of a set of predicate declarations $\mathcal{P}$, a mapping $\mathcal{D}$ of constants to domains, and a set of Knowledge Categories $\mathcal{C}$. The definition of a Knowledge List is given in Definition \ref{theo:kl-def}.

\begin{definition}[Knowledge List]\label{theo:kl-def}\vspace{-0.2cm}
    \begin{equation*}
    \begin{split}
        \mathcal{P} = \{ & P_{1}(D_A), P_{2}(D_B, D_C), \dots, P_{W}(D_X) \}\\
        \mathcal {D} = \{ & D_A \mapsto \{k_{a1}, k_{a2}, \dots, k_{aA} \}\\
                & \cdots \\
                & D_X \mapsto \{k_{x1}, k_{x2}, \dots, k_{xC} \} \}\\
        \mathcal{C} = \{ & c_1, c_2, c_3, \dots, c_{Y}\}
    \end{split}
    \end{equation*}
where $W$ is the number of predicate declarations, $X$ the number of domains in the predicate declarations, $A$, $B$ and $C$ the number of constants $k$ in each respective domain and $Y$ the number of Knowledge Categories in the Knowledge List.\qed
\end{definition}

Building and maintaining a Knowledge List is required to keep track of all known knowledge elements. In MLN-CLA, the initial Knowledge List is constantly updated by merging it with another Knowledge List created from MLNs trained on the new evidence as illustrated in \autoref{fig:mln-cla-overview} on the right. When new evidence is fed into MLN-CLA, first all elements, i.e. predicates, domains and constants, are compared against the Knowledge List. Elements not in the list are marked as unknown and are added to the list after training. Elements already in the list are marked as known and used for training and evidence counting. During the MLN-CLA process the Knowledge List is updated according to the new evidence. Only the elements in the Knowledge List that are affected by the new evidence will be updated. This method enables selective updating of knowledge and incremental learning of new knowledge.

Example \ref{ex:kl-ex} illustrates an example of a Knowledge List with several predicate declarations and domain to constant mappings. In this example the three predicate declarations all share a domain. Each predicate defines a relation between an \texttt{object} and another concept, i.e. its \texttt{size} or \texttt{weight}. Intuitively, because these predicates share domains, they define knowledge within the same context. This property of the predicate and domains declaration is exploited in the Knowledge Categories to structure the information in the Knowledge List.


\newtheorem{example_}{Example}
\begin{example_}\label{ex:kl-ex}\vspace{-0.2cm}
\begin{equation*}
    \begin{aligned}
    \mathcal{P} & = \{Shape(object, shape), Affordance(object, action) \}\\
    \mathcal {D} & = \{
                \quad object \mapsto \{ Ball, Glass, Shoe, Chair, \dots\}\\
                & \qquad \quad shape \mapsto \{Round, Sphere, Cylinder, Cube, \dots\}\\
                & \qquad \quad action \mapsto \{Push, Throw, Pull, Open, \dots\}
                \}\\
    \mathcal{C} & = \{ C_1, C_2, C_3, \dots, C_N\}\\
    \end{aligned}
\end{equation*}
\end{example_}

\subsection{Updating Knowledge Categories}

The second form of knowledge container is the Knowledge Category. A Knowledge Category $C$ consist of an index or identifier $i$, a set of triplets $(F_j, w_j, z_j)$ of associated formulas $F_j$, weights $w_j$ and formula evidence counts $z_j$, and a set of category domains $\mathcal{D}_i$. A Knowledge Category contains a finite number of knowledge triplets and domains. All knowledge triplets in a Knowledge Category contain formulas on the same concepts as specified by the domains. The definition of a Knowledge Category is given in Definition \ref{theo:kc-def}. Knowledge Categories are always part of a Knowledge List as they share the same context. The index $i$ is used to keep track of the Knowledge Category within a Knowledge List over time.

\begin{definition}[Knowledge Category]\label{theo:kc-def}\vspace{-0.2cm}
    \begin{equation*}
        \begin{aligned}
            \textrm{index} = & \; i \\
            (F, w, z) = & \{(F_1, w_1, z_1), (F_2, w_2, z_2), \dots, (F_N, w_N, z_N)\}\\
            \mathcal{D}_i = & \{D_{A}, D_{B}, \dots, D_{X}\}\\
        \end{aligned}
    \end{equation*}
    where $i$ is an integer index, $N$ is the number of knowledge triplets in the category, and $X$ is the number of domains in the formulas of the category. \qed
\end{definition}

Example \ref{ex:kc-ex} shows a possible Knowledge Category within the Knowledge List from Example \ref{ex:kl-ex}. It contains two formulas relating the size of an object to possible actions. The constants $\texttt{Large}$, $\texttt{Push}$, $\texttt{Small}$ and $\texttt{Throw}$ are mapped to their respective domains $\texttt{object}$, $\texttt{size}$ and $\texttt{action}$ via the Knowledge List. Each formula in the knowledge triplets is accompanied by its weight and evidence count. The evidence counts suggest these formulas were learned from different evidence databases as the count is only based on predicate occurrence in the data.


\begin{example_}\label{ex:kc-ex}\vspace{-0.2cm}
    \begin{equation*}
        \begin{aligned}
            \textrm{index} = & \; 3 \\
            (F, w, z) = &\\
            \{(Size(o, Large) \implies & Affordance(o, Push), 0.563, 8),\\
            (Size(o, Small) \implies & !Affordance(o, Throw), -1.27, 4)\}\\
            \mathcal{D}_3 = & \{\texttt{object}, \texttt{size}, \texttt{action}\}\\
        \end{aligned}
    \end{equation*}
\end{example_}


A Knowledge Category of a Knowledge List is unique within the Knowledge List. It contains all formulas in a Knowledge List that relate to the same concept, essentially forming a knowledge cluster. Knowledge Categories are initially created from one formula. A Knowledge Category can contain multiple different formulas. Determining which formula to add to which Knowledge Category is the most important mechanism of MLN-CLA. Any formula in an MLN belongs to only one Knowledge Category in the Knowledge List.

To determine to which Knowledge Category a formula belongs, the algorithm can look at either the predicates or domains of a formula. All predicates and the domains they range over are known through the MLN predicate declarations. A Knowledge Category is made unique by either the set of predicates or domains of the formulas in the category. In practice, there are more relations between domains, as defined by the predicates, than there are domains defined in an MLN. We made the design choice to make Knowledge Categories unique by their domain set, where the set is formed by all domains of each predicate in a formula as described. In MLN-CLA the set of domains $\mathcal{D}_i$ of a Knowledge Category fully determine the category.
This definition enables the merging of Knowledge Categories, resulting in clusters in the knowledge base that can be converted into an MLN for querying or further learning. 



If the domain set of a Knowledge Category $C_1$ is a subset or is equal to the domain set of another Knowledge Category $C_2$, the two Knowledge Categories share common knowledge. All formulas in $C_1$ define rules over the same knowledge domains as $C_2$. The formulas and constants of the categories can be merged together into one Knowledge Category, that is
$C_2(\{F_1, w_1, z_1\} \cup \{F_2, w_2, z_2\}, D_2) \iff D1 \subseteq D2 \wedge C_1(\{F_1, w_1, z_1\}, D_1) \cup C_2(\{F_2, w_2, z_2\}, D_2)$.

The Knowledge Category merging operation consists of two sub-operations: a domain set union and merging the knowledge triplets from one category into the other. During the knowledge triplet merging operation, each formula in the triplets is checked for whether it is new to the category or it is the same as an existing formula in the category. In the case that a formula is new, i.e. it is not the exactly same as an existing one, its triplet is simply added to the category. In the case that a formula is the same, a conflict situation occurs. Two triplets from two different categories each with the exact same formula cannot simply be merged together. The weights of the formulas are learned from different sets of evidence. Even when the weights are exactly the same, the evidence they are based on could differ and thus the knowledge encoded in the weights is different. One of the formula weights must be accepted to merge the two categories. 

\textbf{Step 1} -- Because the domains of two categories that should be merged are a sub/super set of each other, the union of the two domain sets results in the superset. The constants the domains of both categories range over, are assigned to the corresponding domains in the Knowledge List.

\textbf{Step 2} -- To solve formula weight conflicts, it is assumed that formula weights in conflicts always represent different sets of evidence. This assumption allows the formulation of knowledge updating strategies to solve formula weight merging conflicts: the \textit{CL-Naive}, \textit{CL-Conservative} and \textit{CL-Balanced} strategies. The pseudocode for each strategy is shown in \autoref{tab:strategies}. 

\begin{table}[bp]
    \caption{Pseudocode of three knowledge updating strategies. $w$ is the weight of a formula, $z$ is the evidence count.} 
    \centering
    \begin{tabular}{c|c|c}
        \textbf{CL-Naive} & \textbf{CL-Conservative} & \textbf{CL-Balanced}\\ \hline
         $w_1 \gets w_2$ & $ if(z_2 > z_1):$ & $if (z_1 = z_2 = 0):$ \\ 
         $z_1 \gets z_2$ &  $ w_1 \gets w_2 $ & $w1 \gets \frac{w_1+w_2}{2}$, else \\
         & $z_1 \gets z_2 $ & $w_1 \gets \frac{z_1 w_1 + z_2 w_2}{z_1+z_2}$,\;
    $z_1 \gets z_1 + z_2$\\
    \end{tabular}
    \label{tab:strategies}
\end{table}

\textbf{CL Naive} -- The simplest possible strategy is to prefer any new knowledge over the old knowledge. This is represented in the CL-Naive strategy, described in  \autoref{tab:strategies}. In this strategy, conflicts in merging knowledge triplets $(F, w, z)$ with the same formula $(F_1 = F_2)$ are always resolved by copying the new triplet over the old triplet, regardless of how much evidence either of the formula weights was trained on. That the weights are overwritten after each learning step using this strategy, does not mean all previously learned information is lost. During conversion of a Knowledge Category to an MLN, the formula weights are set to their respective current weights prior to learning. Thus each weight is adjusted based on the new evidence from the starting point of the weight from the previous learning step, i.e. $w_{t+1} = w_{t} + \Delta w$ where $\Delta w$ is the newly learned weight difference. This way all learned information is carried over between learning steps.

\textbf{CL Conservative} -- A more conservative strategy is to only accept new knowledge as better if it is based on more evidence than the old knowledge. This is the CL-Conservative strategy, described in \autoref{tab:strategies}. In this strategy the evidence count portion of the knowledge triplets is important. The evidence count of a formula simply represents the amount of evidence seen during the training of the weight of the formula. Only the evidence corresponding to the predicates in the formula are counted as only they contribute to the weight during training. If the evidence count of the new knowledge is larger than that of the old then the new knowledge triplet overwrites the old. In subsequent learning steps the new evidence count is then the barrier to overcome. The old knowledge triplet is not overwritten if the evidence count for the new knowledge is equal or smaller than the old count. 

In contrast to the CL-Naive strategy, the CL-Conservative strategy only updates weights of a formula if it is a new formula or there is more evidence for the weight of the formula than in the previous step. Because new evidence counts overwrite old evidence counts in this strategy, in the next step the new evidence is counted from zero. Another approach is to start counting the new evidence from the previous evidence count to force new formula weights to contain more information than the previous weights.

\textbf{CL Balanced} -- The CL-Balanced strategy, described in \autoref{tab:strategies}, applies this continuous counting approach. To solve same formula merging conflicts the balanced strategy takes the weighted average of the old and new formula weights based on their respective evidence counts. The resulting averaged formula weight contains partial information of each weight in proportion to their respective evidence. After calculating the formula weight average, the evidence counts of both knowledge triplets are summed. The sum of the evidence counts represents that the formula weights were combined and not overwritten. 

If both evidence counts are zero the arithmetic average of the two weights is taken to prevent division by zero. This situation only occurs if two Knowledge Categories are merged with untrained formulas or formulas that have not yet seen any evidence. New evidence can contain only predicate declarations and formulas of yet unknown predicates and domains without any further supporting evidence. In this case Knowledge Categories, with knowledge triplets with weight zero and evidence count zero, are created for each predicate declaration. If any of these categories contain domain subsets of other newly created categories, they are subsequently merged together with the arithmetic average of the zero weights. The zero weight of the original formula is thus conserved. In normal operation this scenario is unlikely to happen. 
As the number of learning steps grows, the evidence count of the CL-Balanced strategy will grow too. Except for the case where there is no new evidence for a formula, then the evidence count is zero and the corresponding weight will not contribute anything to the weighted average. Due to the evidence count growth, the influence of new evidence will have less impact on the merged formula weight if its evidence count remains similar to the previous step. In the CL-Balanced strategy the number of evidence must grow in proportion to the accumulated evidence of previous steps to prevent formula weight from converging. \vspace{-0.2cm}

\subsection{Learning process}
In the first step an MLN and an evidence database are given as initial input to create a knowledge base. In the next time step a new MLN, evidence database or both are introduced to the knowledge base. The new information is compared against the knowledge base to determine for each piece of information whether its is known or unknown. If all information is known then structure learning is performed using the standard Alchemy algorithm \cite{kok2005mln-struct-learning}. If any evidence is unknown only formula weight learning \cite{singla2005discriminative} is performed. The structure learning threshold can be adjusted according to user preference.

After comparing against the knowledge base, each piece of information is put into a Knowledge Category. Only the Knowledge Categories changed by the new information are converted to independent MLNs. Subsequently, if structure learning is performed then the MLN splits are first merged together before structure learning on the new information. Otherwise, the MLN splits' formula weights are trained on the new evidence. Afterwards, the splits are merged together and a new knowledge base is created from the merged or structure learned MLN. Finally, the new knowledge base is merged with the initial knowledge base to incorporate the newly learnt formula weights or structure. In this final merging process the knowledge updating strategies are applied to solve conflicts between old and new formula weights for all duplicate formulas. 
For the cumulative learning of an MLN, new evidence consists of one or a combination of: new formulas (from an MLN), new evidence (from a database), or new predicate declarations that define new relations over either existing or new domains of constants (from either an MLN or database).

\subsection{Cumulative Learning Algorithm}
As indicated in \autoref{fig:mln-cla-overview}, MLN-CLA allows for two types of sources of new evidence: either from an MLN or an evidence database. At each time step a combination of any type of evidence can be presented to the cumulative learner to incorporate. The Alchemy software \cite{kok2005alchemy} used for MLN learning only requires predicates used in formulas in any MLN to be declared beforehand. Within MLN-CLA this requirement is extended to evidence databases. Predicates in evidence must also be declared to indicate to which domains the arguments of evidence predicates belong. For example: \texttt{IsA(object, category)} is a predicate declaration that defines the `IsA' relation over the domains `object' and `category'. The domains are filled with constants from the evidence databases such as \texttt{IsA(Cat, Animal)}, indicating by argument index that `Cat' belongs to the `object' domain and `Animal' belongs to the `category' domain. This assumption enables the cumulative learner to incorporate new relations over new or existing domains in the knowledge base. Structure learning can be applied to newly discovered predicate declarations to construct formulas from these and other predicates.
%
MLN-CLA employs the Knowledge List concept to manage a knowledge base. First-order logic formulas are clustered together within the Knowledge List in Knowledge Categories. These categories can be converted to standalone MLNs to update weights, add new formulas or modify the network structure. After the update, the MLN is converted back to a standalone Knowledge List. Each standalone Knowledge List is then merged into the original Knowledge List using a knowledge updating strategy to solve conflicts. Subsequently, the algorithm is ready to accept new information. 



\section{Experiments}\label{sec:experiments}
We conducted two experiments with MLN-CLA to test the learning capacities in different scenarios. In the first experiment the MLN model is shown new constants (objects such as \texttt{Horse} or \texttt{Small\_boat}) in each learning step whilst keeping the model formulas unchanged. In the second experiment a new formula is introduced to the model in each learning step. In both experiments the three knowledge updating strategies CL-Naive, CL-Conservative and CL-Balanced are tested against two baseline methods.

The MLN model used in the experiments is based on Zhu et al. \cite{zhu2014kbreasoning}. The MLN consists of five formulas as shown in \autoref{tab:mln-formulas}. The `+' symbol indicates that a separate formula weight is learned for each constant that the variable ranges over. The resulting trained model contains a formula with learned weight for each combination of possible constants of the variables preceded by `+' in the untrained MLN. The formulas 1-4 relate an object property to an affordance. These formulas represent a zero-shot approach to object affordance inference as objects are not learned by their label but from their properties \cite{xian2018zero}. Formula 5 relates object categories (such as \texttt{Writing\_implement} and \texttt{Animal}) to each other. This formula acts as supporting information for formula 1. It is a Markov logic representation of an ontology for object categories. The constants and formulas used for training each model are based on the training set of the FOL affordance dataset by Zhu et al. \cite{zhu2014kbreasoning}. 

\begin{table*}[tp]
    \centering
    \caption{MLN model for object affordance inference based on Zhu et al \cite{zhu2014kbreasoning}.}\vspace{-0.2cm}
    \label{tab:mln-formulas}
    \begin{tabular}{>{\centering\arraybackslash}p{0.06\linewidth}|l}
         \textbf{Index}&  \textbf{Formula}\\
         \hline
         1&  $\texttt{IsA(obj,+category)} \Rightarrow \texttt{HasAffordance(obj,+affordance)}$\\
         2&  $\texttt{HasVisualAttribute(obj,+attribute)} \Rightarrow \texttt{HasAffordance(obj,+affordance)}$\\
         3&  $\texttt{HasWeight(obj,+weight)} \Rightarrow \texttt{HasAffordance(obj,+affordance)}$\\
         4&  $\texttt{HasSize(obj,+size)} \Rightarrow \texttt{HasAffordance(obj,+affordance)}$\\
         5&  $\texttt{IsA(obj,+category)} \Rightarrow \texttt{IsA(obj,+category)}$\\
    \end{tabular}
    \vspace{-0.3cm}
\end{table*}

The training set evidence consists of evidence for 5 predicates, such as \texttt{HasAffordance} or \texttt{HasWeight}, and 40 different objects, such as \texttt{Banjo}, \texttt{Cat} or \texttt{W1} where \texttt{W1} is a constant representing the weight class 'less than 1 kilogram'. The model performance of each learning step of each experiment is evaluated on the test dataset by Zhu et al. \cite{zhu2014kbreasoning}. The test set consists of the evidence for the same 5 predicates for 22 objects. The test objects are similar to the training set but not the same.

In each experiment, the trained MLN at the end of a learning step is queried for the \texttt{HasAffordance} predicate given all test evidence. The model outputs the predicted marginal probability -- between 0 (false), 0.5 (unknown) and 1 (true) -- that an object has a certain affordance. To evaluate the performance we adopt the Area under the ROC curve (AUC) metric, similar to Richardson and Domingos \cite{richardson2006mln}. In each experiment the baseline performance is set by the standard weight learning algorithms in Alchemy called the \textit{generative} \cite{richardson2006mln} and \textit{discriminatively} \cite{singla2005discriminative} learned MLN models, both trained on the whole training set including all formulas.


\subsection{Learning new constants}
In \autoref{fig:result-cl-const} the average performance of the constants learning experiment is shown over 300 runs. In each run the order of the evidence fed to MLN-CLA was randomised. We expected MLN-CLA to approach but not outperform the baseline batch trained MLNs, indicated in Figures \ref{fig:result-cl-const}-\ref{fig:results-cl-formulas} with \textit{Generative} \cite{richardson2006mln} and \textit{Discriminative} \cite{singla2005discriminative}. Initially the MLN-CLA variants do perform similar to the baseline MLNs. However, from 50\% of the object constants evidence seen, CL-Naive and CL-Conservative decline in performance or drop below the baselines. Surprisingly, the CL-Balanced variant outperforms the baselines from the second learning step. We hypothesise that the CL-Balanced strategy acts as a regulariser on the weights. A regulariser affects the weights by making formulas that have a larger impact on performance, e.g. $\texttt{HasWeight(object, Heavy)} \Rightarrow \texttt{HasAffordance(object, Lift)}$, have larger weights (either negative or positive). It pushes the weights for less important formulas, e.g. $\texttt{HasSize(obj, Small)} \Rightarrow \texttt{HasAffordance(obj, Row)}$, to zero.

\subsection{Learning new formulas}
In \autoref{fig:results-cl-formulas} the average performance of the formulas learning experiment is shown over 5 runs. In each run the order of formulas learned is different to ensure that the results are invariant to evidence order. In this experiment the cumulative learner learns a new predicate in each step. However, the test set contains evidence of predicates not yet known to the cumulative learner. As a consequence, MLNs trained on only a subset of the predicates present in the evidence cannot be evaluated on the full test set. We circumvented this problem by only evaluating on the test evidence of which the predicates are known to the incremental learning. Consequently, the results in \autoref{fig:results-cl-formulas} are only comparable within one step and not between steps. This is indicated in the figure by the lack of a connecting line between each data point. 
In the figure, the MLN-CLA variants have overlapping performance for the first three learning steps. In the fourth and fifth learning steps the MLN-CLA variants showcase their differences. The CL-Conservative outperforms both the baselines and the other variants after having seen all five formulas. In the same learning step, CL-Balanced performs the worst of the three MLN-CLA variants, although it still outperforms the baselines.

\subsection{Knowledge updating strategies}
\begin{figure*}[!h]
    \centering
    \begin{subfigure}{0.485\linewidth}
        \centering
        \includegraphics[width=\linewidth]{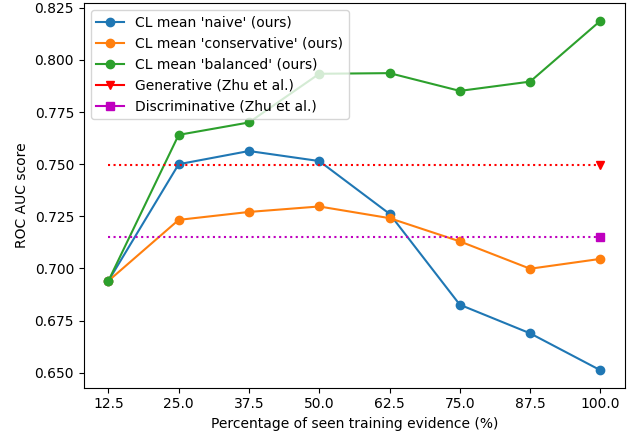}
        \vspace{-0.7cm}
        \caption{Learning new \texttt{object} constants.}
        \label{fig:result-cl-const}
    \end{subfigure}
    \begin{subfigure}{0.47\linewidth}
        \centering
        \includegraphics[width=\textwidth]{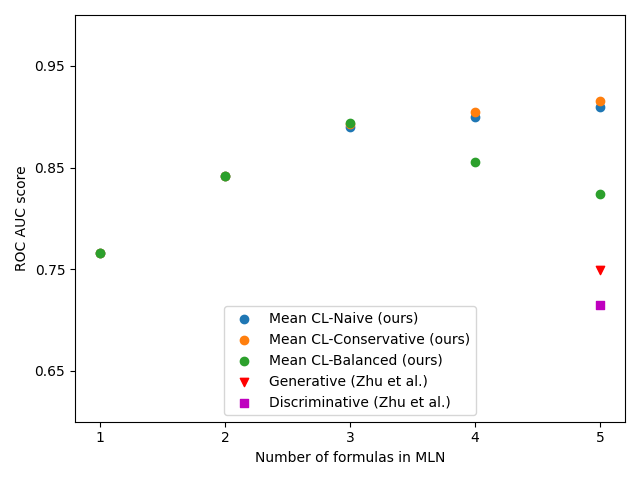}
        \vspace{-0.7cm}
        \caption{Learning new formulas.}
        \label{fig:results-cl-formulas}
    \end{subfigure}\vspace{-0.1cm}
    \caption{Learning of new constants and new formulas.}
\end{figure*}

\begin{figure*}[!h]
    \centering
    \includegraphics[width=\linewidth]{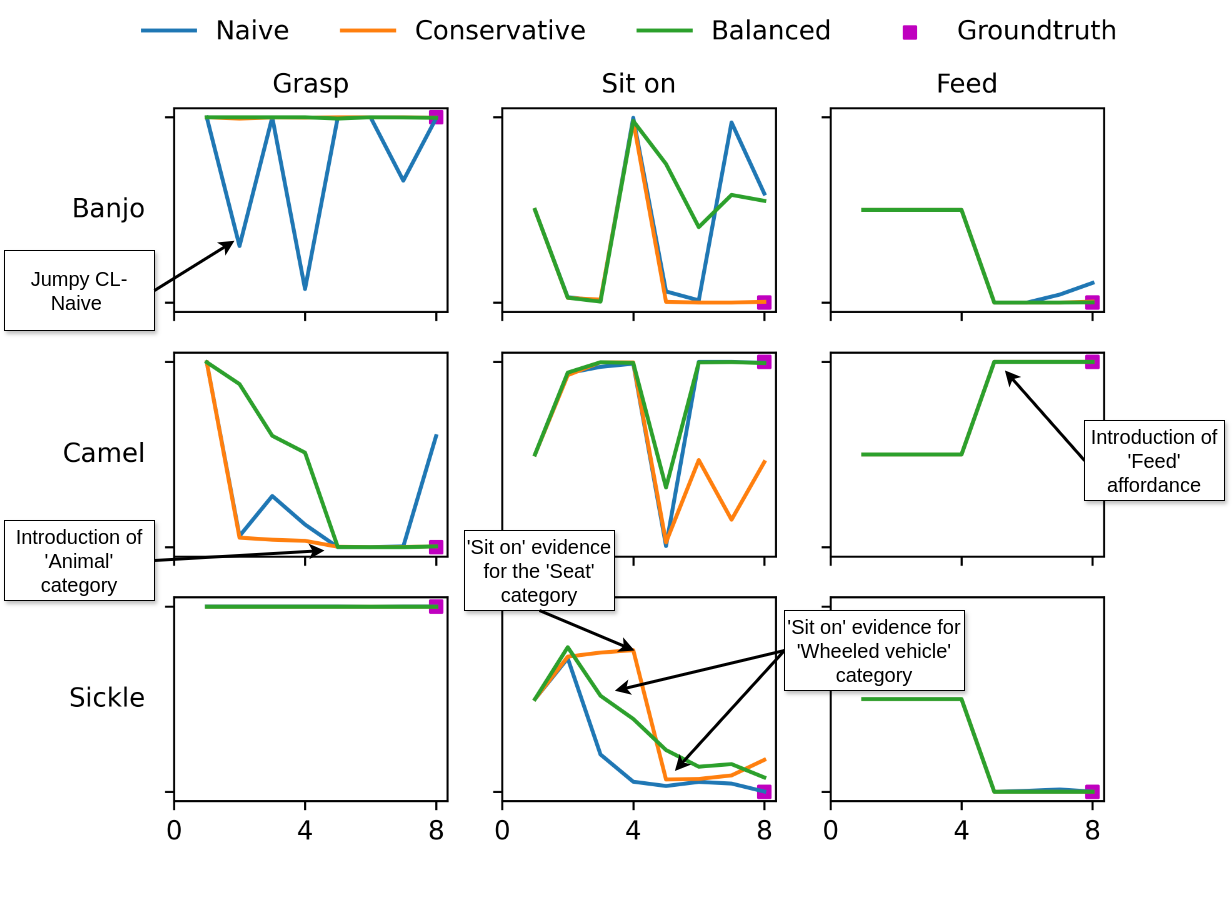}
    \vspace{-0.5cm}
    \caption{Marginal probability predictions of the three knowledge updating strategies over the steps in the constants learning experiment.}\vspace{-0.4cm}
    \label{fig:result-marg-prob-selection}
\end{figure*}

In Figures \ref{fig:result-cl-const} and \ref{fig:results-cl-formulas} the knowledge updating strategies show different behaviours under different scenarios. In the constant learning setting the CL-Balanced strategy performs best, whereas in the formula learning setting it performs worst of the three strategies. There seems to be a trade-off between when each strategy can be applied best. The differences between the strategies are highlighted by \autoref{fig:result-marg-prob-selection}. Comparison of strategies is made difficult by the fact that the effects on formula weights by each strategy in earlier step propagate to subsequent steps. In the figure the predicted marginal probabilities for nine object-affordance pairs over the eight learning steps of the constants learning experiment are shown. Each knowledge updating strategy reacts differently to new evidence. CL-Naive adjusts weights every time new evidence comes in. A good example of this behaviour is shown in the top-left in \autoref{fig:result-marg-prob-selection}. The CL-Conservative strategy is a less erratic version of CL-Naive. CL-Balanced is even less erratic and introduces a delay to a flip in predictions such as shown in the pair Camel-Grasp. In summary, there is no one strategy that performs best. However, the cumulative learning strategies do outperform the standard MLN batch learning algorithms \cite{richardson2006mln, singla2005discriminative}. This observation invites more research into the knowledge updating strategies. Other strategies than the three presented here are possible. 

\section{Conclusions}\label{sec:conclusion}

We introduced MLN-CLA for the incremental learning of new knowledge and applied our algorithm to an affordance learning scenario. To prevent knowledge from being lost during learning we additionally introduced three knowledge updating strategies. Each strategy solves knowledge merging conflicts in a different manner based on evidence counts. Within MLN-CLA the strategies can easily be modified.

Our experiments show the effectiveness of MLN-CLA in adapting to new evidence. 
In addition, the experiments highlight the differences and trade-offs between the choice of knowledge updating strategy to apply. The CL-Balanced strategy performs best overall when only learning new constants. The same strategy performs worst of the three whilst still outperforming the batch learned baselines.

The knowledge updating strategies we formulated rely heavily on the evidence count (except for CL-Naive). In our experiments, we did not vary the batch sizes of new evidence per time step. The impact of varying batch sizes over time on performance and system stability should be further investigated. Our current implementation of MLN-CLA uses the default Alchemy weight and structure learning algorithms. An interesting exercise would be to replace those by the Online Structure Learning algorithm to improve learning performance.

\bibliographystyle{apalike}
\bibliography{bibfile}

\begin{thebibliography}{}

\bibitem[Andries et~al., 2018]{mihai2018aff-eq}
Andries, M., Chavez-Garcia, R.~O., Chatila, R., Giusti, A., and Gambardella, L.~M. (2018).
\newblock Affordance equivalences in robotics: A formalism.
\newblock {\em Frontiers in Neurorobotics}, 12.

\bibitem[Ard{\'o}n et~al., 2020]{ardon2020affordances}
Ard{\'o}n, P., Pairet, {\`E}., Lohan, K.~S., Ramamoorthy, S., and Petrick, R. (2020).
\newblock Affordances in robotic tasks--a survey.
\newblock {\em arXiv preprint arXiv:2004.07400}.

\bibitem[Ben-David et~al., 1997]{benDavid1997offline}
Ben-David, S., Kushilevitz, E., and Mansour, Y. (1997).
\newblock Online learning versus offline learning.
\newblock {\em Machine Learning}, 29(1):45--63.

\bibitem[Cui et~al., 2022]{cui2022mcla}
Cui, S., Zhu, T., Zhang, X., and Ning, H. (2022).
\newblock {MCLA: Research on cumulative learning of Markov Logic Network}.
\newblock {\em Knowledge-Based Systems}, 242:108352.

\bibitem[Domingos and Lowd, 2019]{domingos2019unifying}
Domingos, P. and Lowd, D. (2019).
\newblock Unifying logical and statistical ai with markov logic.
\newblock {\em Commun. ACM}, 62(7):74–83.

\bibitem[French, 1999]{french1999catastrophic}
French, R.~M. (1999).
\newblock Catastrophic forgetting in connectionist networks.
\newblock {\em Trends in Cognitive Sciences}, 3(4):128--135.

\bibitem[Gibson, 1979]{gibson1979}
Gibson, J.~J. (1979).
\newblock {\em The Ecological Approach to Visual Perception}.
\newblock Boston: Houghton Mifflin.

\bibitem[Huynh and Mooney, 2011]{huynh2011osl}
Huynh, T.~N. and Mooney, R.~J. (2011).
\newblock Online structure learning for markov logic networks.
\newblock In Gunopulos, D., Hofmann, T., Malerba, D., and Vazirgiannis, M., editors, {\em Machine Learning and Knowledge Discovery in Databases}, pages 81--96, Berlin, Heidelberg. Springer Berlin Heidelberg.

\bibitem[Jaramillo-Cabrera et~al., 2019]{cabrera2019bayesian}
Jaramillo-Cabrera, E., Morales, E.~F., and Martinez-Carranza, J. (2019).
\newblock Enhancing object, action, and effect recognition using probabilistic affordances.
\newblock {\em Adaptive Behavior}, 27(5):295--306.

\bibitem[Kok and Domingos, 2005]{kok2005mln-struct-learning}
Kok, S. and Domingos, P. (2005).
\newblock Learning the structure of markov logic networks.
\newblock In {\em Proceedings of the 22nd international conference on Machine learning}, pages 441--448.

\bibitem[Kok et~al., 2005]{kok2005alchemy}
Kok, S., Singla, P., Richardson, M., and Domingos, P. (2005).
\newblock {The Alchemy system for statistical relational AI}.
\newblock Technical report, Department of Computer Science and Engineering, University of Washington, Seattle, WA.

\bibitem[Luis~Montesano and Santos-Victor., 2007]{montesano2007}
Luis~Montesano, Manuel~Lopes, A.~B. and Santos-Victor., J. (2007).
\newblock Modeling affordances using bayesian networks.
\newblock {\em IEEE/RSJ International Conference on Intelligent Robots and Systems}, pages 4102--4107.

\bibitem[Pearl, 1988]{pearl1988bp}
Pearl, J. (1988).
\newblock {\em Probabilistic Reasoning in Intelligent Systems: Networks of Plausible Inference}.
\newblock Morgan Kaufmann Publishers Inc., San Francisco, CA, USA.

\bibitem[Richardson and Domingos, 2006]{richardson2006mln}
Richardson, M. and Domingos, P. (2006).
\newblock Markov logic networks.
\newblock {\em Machine Learning}, 62:107--136.
\newblock https://doi.org/10.1007/s10994-006-5833-1.

\bibitem[Singla and Domingos, 2005]{singla2005discriminative}
Singla, P. and Domingos, P. (2005).
\newblock Discriminative training of markov logic networks.
\newblock In {\em Proceedings of the 20th National Conference on Artificial Intelligence - Volume 2}, AAAI'05, page 868–873. AAAI Press.

\bibitem[Th{\'o}risson et~al., 2019]{thórisson2019cumulative}
Th{\'o}risson, K.~R., Bieger, J., Li, X., and Wang, P. (2019).
\newblock Cumulative learning.
\newblock In Hammer, P., Agrawal, P., Goertzel, B., and Ikl{\'e}, M., editors, {\em Artificial General Intelligence}, pages 198--208, Cham. Springer International Publishing.

\bibitem[Xian et~al., 2018]{xian2018zero}
Xian, Y., Lampert, C.~H., Schiele, B., and Akata, Z. (2018).
\newblock Zero-shot learning—a comprehensive evaluation of the good, the bad and the ugly.
\newblock {\em IEEE transactions on pattern analysis and machine intelligence}, 41(9):2251--2265.

\bibitem[Zhu et~al., 2014]{zhu2014kbreasoning}
Zhu, Y., Fathi, A., and Fei-Fei, L. (2014).
\newblock Reasoning about object affordances in a knowledge base representation.
\newblock In Fleet, D., Pajdla, T., Schiele, B., and Tuytelaars, T., editors, {\em Computer Vision -- ECCV 2014}, pages 408--424, Cham. Springer International Publishing.

\end{thebibliography}

\end{document}